\documentclass[a4paper,fleqn]{cas-sc}
\usepackage{amsthm}
\usepackage{booktabs}
\usepackage{array}
\usepackage{hyperref}
\usepackage{overpic}
\usepackage{xcolor}
\usepackage{pifont}

\usepackage{amssymb}
\usepackage{graphicx}
\usepackage{epstopdf}
\usepackage[english]{babel}
\usepackage[numbers]{natbib}
\usepackage{bm}
\usepackage{float}
\usepackage{algorithm}
 \usepackage{dsfont}
\usepackage{multirow}
\usepackage{longtable}
\usepackage{subfigure}
\usepackage{supertabular}
\usepackage{algpseudocode}
\usepackage{amsmath}
\usepackage[justification=centering]{caption}

\usepackage{algorithm,float}

\def\tsc#1{\csdef{#1}{\textsc{\lowercase{#1}}\xspace}}
\tsc{WGM}
\tsc{QE}

\begin{document}
\let\WriteBookmarks\relax
\def\floatpagepagefraction{1}
\def\textpagefraction{.001}

\shorttitle{}    

\shortauthors{M. Jiang et al.}  

\title [mode = title]{Interpretable Fair  Clustering}  



%

\author[1]{Mudi Jiang}[orcid=0000-0001-9474-8375]


\ead{792145962@qq.com}

\author[1]{Jiahui Zhou}



\author[1]{Xinying Liu}



\author[1]{Zengyou He}


\author[1]{Zhikui Chen}
\cormark[1]

\ead{zkchen@dlut.edu.cn}

\affiliation[1]{organization={School of Software, Dalian University of Technology}, 
            city={Dalian},
            state={Liaoning},
            country={China}}


\cortext[1]{Corresponding author}



\begin{abstract}
Fair clustering has gained increasing attention in recent years, especially in applications involving socially sensitive attributes. However, existing fair clustering methods often lack interpretability, limiting their applicability in high-stakes scenarios where understanding the rationale behind clustering decisions is essential. In this work, we address this limitation by proposing an interpretable and fair clustering framework, which integrates fairness constraints into the structure of decision trees. Our approach constructs interpretable decision trees that partition the data while ensuring fair treatment across protected groups. To further enhance the practicality of our framework, we also introduce a variant that requires no fairness hyperparameter tuning, achieved through post-pruning a tree constructed without fairness constraints. Extensive experiments on both real-world and synthetic datasets demonstrate that our method not only delivers competitive clustering performance and improved fairness, but also offers additional advantages such as interpretability and the ability to handle multiple sensitive attributes. These strengths enable our method to perform robustly under complex fairness constraints, opening new possibilities for equitable and transparent clustering.
\end{abstract}


\begin{highlights}
\item IFCT introduces interpretability into fair clustering using decision trees.
\item IFCT-P achieves fairness without the need for manual parameter tuning.
\item First to support mixed features, multiple sensitive attributes with interpretability.
\item Experiments demonstrate strong performance in both clustering quality and fairness.
\end{highlights}

\begin{keywords}
 Fair clustering \sep Interpretability \sep Unsupervised learning \sep Decision tree
\end{keywords}

\maketitle

\section{Introduction}\label{1}
Cluster analysis \cite{oyewole2023data}  is a fundamental task in data mining, widely used for uncovering structure in unlabeled data. Traditional clustering algorithms (e.g., $k$-means) are primarily designed to optimize objectives such as intra-cluster compactness or inter-cluster separation, without considering fairness across different demographic groups. As machine learning is increasingly applied in socially sensitive domains, concerns have emerged regarding biased clustering outcomes that may systematically disadvantage certain populations. In response, fair clustering \cite{chhabra2021overview} has received growing attention in recent years, aiming to incorporate sensitive attribute information as fairness constraints during clustering, while preserving utility.

Existing studies on fair clustering  can be broadly categorized into individual fairness \cite{jung_et_al:LIPIcs.FORC.2020.5} and group fairness \cite{NIPS2017_978fce5b}. The former requires that similar individuals receive similar treatment, relying on a well-defined similarity metric. The latter focuses on preventing systematic disadvantages to any demographic group, often formalized through the notion of Disparate Impact \cite{rutherglen1987disparate}. In this work, we focus on group fairness, which is particularly relevant in real-world settings where equitable treatment is critical.
Group fairness methods are generally divided into pre-processing, in-processing, and post-processing approaches, which modify the input data, algorithm, or clustering results, respectively, to improve fairness. Although these methods have effectively improved fairness in clustering outcomes, they generally do not employ interpretable models to generate clustering results. In sensitive domains such as healthcare or finance, interpretability is vital for ensuring transparency and trust. Without clear explanations for how and why clusters are formed, even fair clustering outcomes may be difficult to validate or adopt in practice.

Interpretable clustering \cite{hu2024interpretable} has received increasing attention for its ability to provide human-understandable explanations of cluster assignments, typically through models such as decision trees \cite{gabidolla2022optimal} or \textit{if-then} rules \cite{balachandran2009interpretable}. However, existing methods in this area typically assume unbiased data and have not yet considered fairness across sensitive groups. This disconnect limits their applicability in real-world scenarios where both fairness and interpretability are essential. Therefore, there is a growing need for clustering approaches that can mitigate bias while preserving interpretability.
\begin{table}[htbp]
\centering
\caption{Comparison of IFCT with representative fair clustering algorithms.}
\label{tab1}
\begin{tabular}{@{}lccccc@{}}
\toprule
Algorithm & Fairlets \cite{NIPS2017_978fce5b} &  Fair Scalable \cite{backurs2019scalable} & FairSC \cite{pmlr-v97-kleindessner19b} & FairDEN \cite{zhao2024dfmvc} & IFCT\\
\midrule
Interpretability & \ding{55} & \ding{55} & \ding{55}&\ding{55}  &\ding{51}\\
Multiple sensitive attributes & \ding{55} &\ding{55} & \ding{55} &\ding{51} &\ding{51} \\
Multiple  sensitive groups (within one attribute) & \ding{55} & \ding{55} &\ding{51} &\ding{51} &\ding{51} \\
Mixed-type features & \ding{55}  & \ding{55} &\ding{55} &\ding{51}&\ding{51} \\
\bottomrule
\end{tabular}
\end{table}

Motivated by the aforementioned observations, we propose the Interpretable Fair Clustering Tree (IFCT), a method that jointly optimizes clustering quality and group fairness. Specifically, IFCT integrates a mixed-type distortion loss and a fairness regularization term into a unified objective, enabling it to naturally handle heterogeneous feature types and multiple sensitive attributes while ensuring fair and interpretable clustering. A comparison with representative fair clustering methods is summarized in Table~\ref{tab1}, highlighting IFCT's unique ability to support both interpretability and complex fairness settings.
Furthermore, to reduce the burden of manually tuning the trade-off between clustering quality and fairness, we introduce IFCT-P, an enhanced variant that decouples the optimization process: it first grows an over-expanded tree by optimizing only the quality objective, and then prunes it using a fairness-guided criterion.

To validate the effectiveness of our proposed method, we conduct experiments on both real-world and synthetic datasets. The results show that our method achieves competitive clustering quality and fairness performance compared to existing fair clustering methods, while maintaining high interpretability.

In summary, the main contributions of this paper are outlined as follows:
\begin{itemize}
    \item We propose the IFCT algorithm and its enhanced variant IFCT-P, which jointly ensure interpretability and group fairness in clustering through a decision-tree-based design.
    \item Our method supports mixed-type features and multiple sensitive attributes by incorporating a mixed-type distortion loss and fairness regularization into a unified optimization objective.
    \item Extensive experiments on real-world and synthetic datasets demonstrate that the proposed methods achieve competitive performance in both clustering quality and  fairness compared to existing baselines, while maintaining  interpretability.
\end{itemize}

The remainder of this paper is organized as follows. Section \ref{2} reviews the related work. Section \ref{3} introduces the proposed framework in detail. Section \ref{4} reports the experimental results and analysis. Finally, Section \ref{5} concludes the paper and outlines directions for future research.









\section{Related work}\label{2}
In this section, we provide a brief overview of related work, focusing on two key areas: fair clustering and interpretable clustering.
\subsection{Fair Clustering}
\subsubsection{Individual fairness}
The concept of individual fairness in clustering was first introduced by Jung et al. \cite{jung_et_al:LIPIcs.FORC.2020.5}, formalizing the idea that each data point should be assigned to a nearby cluster center relative to its local data density. This notion was further extended by Mahabadi et al. \cite{mahabadi2020individual} through a bicriteria approximation framework, allowing controlled trade-offs between clustering cost and fairness guarantees. Subsequent studies \cite{negahbani2021better,pmlr-v151-vakilian22a} have focused on improving approximation bounds under various distance metrics, while others have aimed to enhance the scalability of fairness-aware clustering algorithms for large-scale applications \cite{pmlr-v151-chhaya22a,pmlr-v238-bateni24a}.
\subsubsection{Group fairness}
Group fairness in clustering aims to ensure that different demographic groups are treated equitably throughout the clustering process. According to the stage at which fairness is introduced, existing methods can be divided into three categories: pre-processing methods, in-processing methods, and post-processing methods. 

Pre-processing methods incorporate fairness by transforming or augmenting the input data before clustering. For instance, fairlet decomposition \cite{NIPS2017_978fce5b,NEURIPS2020_f10f2da9} constructs small group-balanced subsets (fairlets) that are then clustered using standard algorithms. Other approaches, such as fair coresets \cite{schmidt2020fair,NEURIPS2019_810dfbbe}, compress the data while preserving fairness constraints, and antidote data \cite{pmlr-v171-chhabra22a} enhances fairness by adding carefully selected samples to the dataset.
In-processing methods enforce fairness during clustering by modifying the algorithm or its objective function. Some approaches integrate fairness constraints directly into traditional formulations like spectral or $k$-median clustering \cite{kleindessner2019guarantees,10.1007/978-3-030-86520-7_47}, while others adopt adversarial learning or multi-objective optimization within deep clustering frameworks to jointly optimize for clustering quality and group fairness \cite{zhang2021deep,li2020deep}.
Post-processing methods address fairness after clustering has been performed. These techniques do not alter the clustering objective itself but instead adjust cluster assignments or centers to satisfy group fairness requirements \cite{kleindessner2019fair,jones2020fair}.

Most of these methods focus on achieving fairness while maintaining clustering loss within a bounded approximation of the optimal solution, thereby quantifying the trade-off often referred to as the “price of fairness.” However, to better preserve fairness in  decision trees for practical applications and to naturally accommodate complex scenarios such as multiple sensitive attributes, we adopt a simple fairness regularization approach, similar in spirit to that employed in \cite{krieger2025fairden,zhao2024dfmvc}.
\subsection{Interpretable Clustering}
In recent years, interpretable models applied to the field of clustering have included, but are not limited to, decision trees \cite{lundberg2020local}, \textit{if-then} rules \cite{balachandran2009interpretable}, convex polyhedral \cite{chen2016interpretable}, and prototypes \cite{davidson2024exemplars}. As this work adopts binary decision trees as the interpretable model for fair clustering, relevant literature is discussed in the following subsections.

\subsubsection{Two-stage Tree Construction}
Two-stage approaches for constructing interpretable clustering trees generally begin by applying standard clustering algorithms to generate pseudo-labels, which are then used to train a supervised decision tree \cite{bandyapadhyay2023find}. The strategy for determining the splitting criterion at each internal node varies across methods. For example, the method proposed in \cite{dasgupta2020explainable} seeks to minimize node misclassification at each split, while the approach in \cite{makarychev2022explainable} computes the median of all cluster centers within a node and selects the split that maximizes the distance from this median. These methods often heavily rely on the quality of the pseudo-labels generated in the first stage. In contrast, Gabidolla et al. \cite{gabidolla2022optimal} introduces a joint optimization framework that iteratively refines both the clustering variables (e.g., centroids in 
$k$-means) and the decision tree parameters, aiming for a more integrated learning process.
\subsubsection{One-stage Tree Construction}
Early decision tree-based clustering methods include the one-stage approach in \cite{liu2000clustering}, which introduces uniformly distributed synthetic data as auxiliary inputs to build a supervised tree by modifying standard splitting criteria such as information gain to separate original and synthetic data. Although this method produces binary splits that are relatively easy to interpret, its dependence on generated data introduces additional assumptions that limit interpretability. Other methods directly utilize the original data without synthetic augmentation. For example, Basak et al. \cite{basak2005interpretable} proposes four distinct criteria for selecting the most informative attribute at each internal node, allowing the tree to capture meaningful structure in an unsupervised manner. In a similar vein, some approaches focus on reducing within-node heterogeneity by selecting split variables and points based on empirical estimates of probabilities and deviations, thereby improving clustering accuracy and coherence  \cite{fraiman2013interpretable,ghattas2017clustering}.
\section{Method}
\label{3}
\subsection{Notations}
Let $\mathcal{X}=\{x_i\}_{i=1}^n$ denote the dataset, where each instance is represented by $d_\text{num}$ numerical features and $d_\text{cat}$ categorical features. Each sample $x_i$ is further associated with a set of sensitive attributes (e.g., race, gender), denoted as $\mathbf{s}_i = [s_i^{(1)}, s_i^{(2)}, \ldots, s_i^{(U)}]$, where $s_i^{(u)} \in \mathcal{S}_u$ and $|\mathcal{S}_u|$ denotes the number of possible groups for the $u$-th attribute.

The objective of interpretable fair clustering is to partition $\mathcal{X}$ into $k$ clusters such that:
\begin{itemize}
\item Samples within each cluster exhibit high similarity in the feature space.
\item The partition can be represented and interpreted by an interpretable model.
\item The clustering results ensure group fairness with respect to sensitive attributes.
\end{itemize}

In this work, we address all three objectives by performing clustering through decision tree construction. 
The objective function and optimization strategy are introduced in the following sections.
\subsection{Optimization Model}
We formalize interpretable fair clustering as an optimization problem that jointly 
minimizes intra-cluster distortion and fairness deviation under a tree-based partition structure. Specifically, a decision tree $\mathcal{T}$ is constructed to hierarchically divide the dataset $\mathcal{X}$ into $k$ disjoint leaf nodes, each corresponding to a cluster defined by an interpretable rule path.
\subsubsection{Overall objective function}
Let $\mathcal{L}_C(\mathcal{T})$ denote the intra-cluster compactness loss
and $\mathcal{L}_F(\mathcal{T})$ the fairness regularization term.
The overall optimization objective is written as:
\begin{equation}
\label{eq:objective}
\min_{\mathcal{T}} \;
\mathcal{L}(\mathcal{T})
= 
\mathcal{L}_C(\mathcal{T})
+ \lambda \, \mathcal{L}_F(\mathcal{T})
=\sum_{v \in \mathbb{L}(\mathcal{T})}
\Big[
\mathcal{L}_C(v)
+ \lambda \, \mathcal{L}_F(v)
\Big],
\end{equation}
where $\mathbb{L}(\mathcal{T})$ denotes the set of leaf nodes in the tree $\mathcal{T}$,
$\mathcal{L}_C(v)$ and $\mathcal{L}_F(v)$ correspond to the compactness and fairness losses of node $v$, respectively, and $\lambda \geq 0$ serves as a trade-off parameter balancing the two losses.

\subsubsection{Intra-cluster compactness}
To effectively model mixed-type data, we introduce a hybrid compactness loss based on the Sum of Squared Errors (SSE) to evaluate intra-cluster similarity. The compactness loss of each node (or cluster) comprises two components corresponding to the numerical and categorical feature spaces.

For numerical features, we employ the standard SSE formulation:
\begin{equation}
\label{eq:num-sse}
\mathcal{L}_{n}(v)
= \sum_{x \in v} \big\| x - \bar{x}_v \big\|_2^2,
\end{equation}
where  $\bar{x}_v$ denotes the mean vector of node $v$.

For categorical features, compactness is quantified using a mode reconstruction loss, which penalizes deviations from the dominant category in each feature.
Let $n_{v,j,g}$ denote the number of samples in node $v$ that take category $g$ for feature $j$. The categorical distortion is then defined as:
\begin{equation}
\label{eq:cat-mode-loss}
\mathcal{L}_{c}(v)
= \sum_{j=1}^{d_{c}}
\Big(n_v - \max_{g} n_{v,j,g}\Big),
\end{equation}
where $n_v$ is the total number of samples in node $v$.
This loss achieves its minimum when all samples share the same category per feature and increases with categorical diversity, effectively reflecting intra-cluster impurity.

The overall intra-cluster compactness loss is computed by aggregating the mixed-type losses across all leaf nodes:
\begin{equation}
\label{eq:tree-loss}
\mathcal{L}_{C}(\mathcal{T}) = \sum_{v \in \mathbb{L} (\mathcal{T})}\mathcal{L}_{C} (v) 
= \sum_{v \in \mathbb{L}(\mathcal{T})}
\Big[
\mathcal{L}_{n}(v)
+ 
\frac{(1-\rho)\,\mathcal{L}_{n}(\mathcal{X})}
     {\rho\,\mathcal{L}_{c}{(\mathcal{X})}+\varepsilon}
\,\mathcal{L}_{c}(v)
\Big],
\end{equation}
where $\rho = \tfrac{d_{n}}{d_{n}+d_{c}}$ denotes the  proportion of numerical features, and $\varepsilon$ is a small constant added to avoid division by zero.
This adaptive weighting balances the influence of numerical and categorical distortions, allowing the compactness objective to fairly account for heterogeneous feature types.
\subsubsection{Fairness regularization}
To mitigate bias across sensitive groups, we incorporate a fairness regularization term that penalizes deviations between the sensitive group distribution within each node and the corresponding global distribution.

Let $G^{(u)}_{v}$ and $G^{(u)}$  denote the empirical and global distributions 
of the $u$-th sensitive attribute with $|\mathcal{S}^{(u)}|$ groups. The fairness deviation of node $v$ is defined as:
\begin{equation}
\label{eq:fair-node}
\mathcal{L}_{F}(v)
= \sum_{u=1}^{U} 
w^{(u)} 
\, \big\| G^{(u)}_{v} - G^{(u)} \big\|_{1}, 
\text{s.t.} \; 
\sum_{u=1}^{U} w^{(u)} = 1, \;  w^{(u)} \ge 0,
\end{equation}
where the weights $\{w^{(u)}\}$ control the relative 
importance of each sensitive attribute, typically set to $1/U$ for equal weighting. This formulation naturally supports multiple sensitive attributes  and ensures balanced fairness optimization across them.

The overall fairness loss of the tree is aggregated over all leaf nodes:
\begin{equation}
\label{eq:fair-tree}
\mathcal{L}_{F}(\mathcal{T})
= \sum_{v \in \mathbb{L}(\mathcal{T})}
\mathcal{L}_{F}(v).
\end{equation}

\subsection{Interpretable Fair Clustering Tree }
\begin{algorithm}[htbp]
\caption{IFCT}
\label{alg:iftree}
\begin{algorithmic}[1]
\Require Dataset $\mathcal{X}$, number of clusters $k$, fairness weight $\lambda$.
\Ensure Interpretable decision tree $\mathcal{T}$ with $k$ leaves.
\State Compute global sensitive group distribution $G$ 
\State Initialize tree $\mathcal{T}$ with root node $r \gets node(\mathcal{X})$
\State Initialize leaf set $\mathbb{L} \gets \{r\}$
\State $\Delta(r)$, $SR_r\gets\textsc{BestRule}(r,\lambda,G)$
\While{$|\mathbb{L}| < k$ }
    \State  $D \gets \arg\max_{v\in\mathbb{L}} \Delta(v)$ 
    \State $(D_L, D_R) \gets {SR_D}(D)$
    \State  Attach $(D_L, D_R)$ as children of $D$ in $\mathcal{T}$
    \State $\mathbb{L} \gets \mathbb{L} \setminus \{D\} \cup \{D_L,D_R\}$
    \State $\Delta(D_L)$, $SR_{D_L}\gets\textsc{BestRule}(D_L,\lambda,G)$
    \State $\Delta(D_R)$, $SR_{D_R}\gets\textsc{BestRule}(D_R,\lambda,G)$
\EndWhile
\State \Return $\mathcal{T}$
\end{algorithmic}
\end{algorithm}
In this section, we present the Interpretable Fair Clustering Tree (IFCT) algorithm.
Jointly optimizing clustering compactness and fairness results in a highly combinatorial, non-convex problem, making global optimization computationally infeasible.
To address this, IFCT adopts a greedy best-first growth strategy, as outlined in Algorithm~\ref{alg:iftree}, which incrementally expands the tree by selecting the split that maximizes local improvement in the overall objective function. 

\begin{figure}[pos=H]
	\includegraphics[scale=0.5]{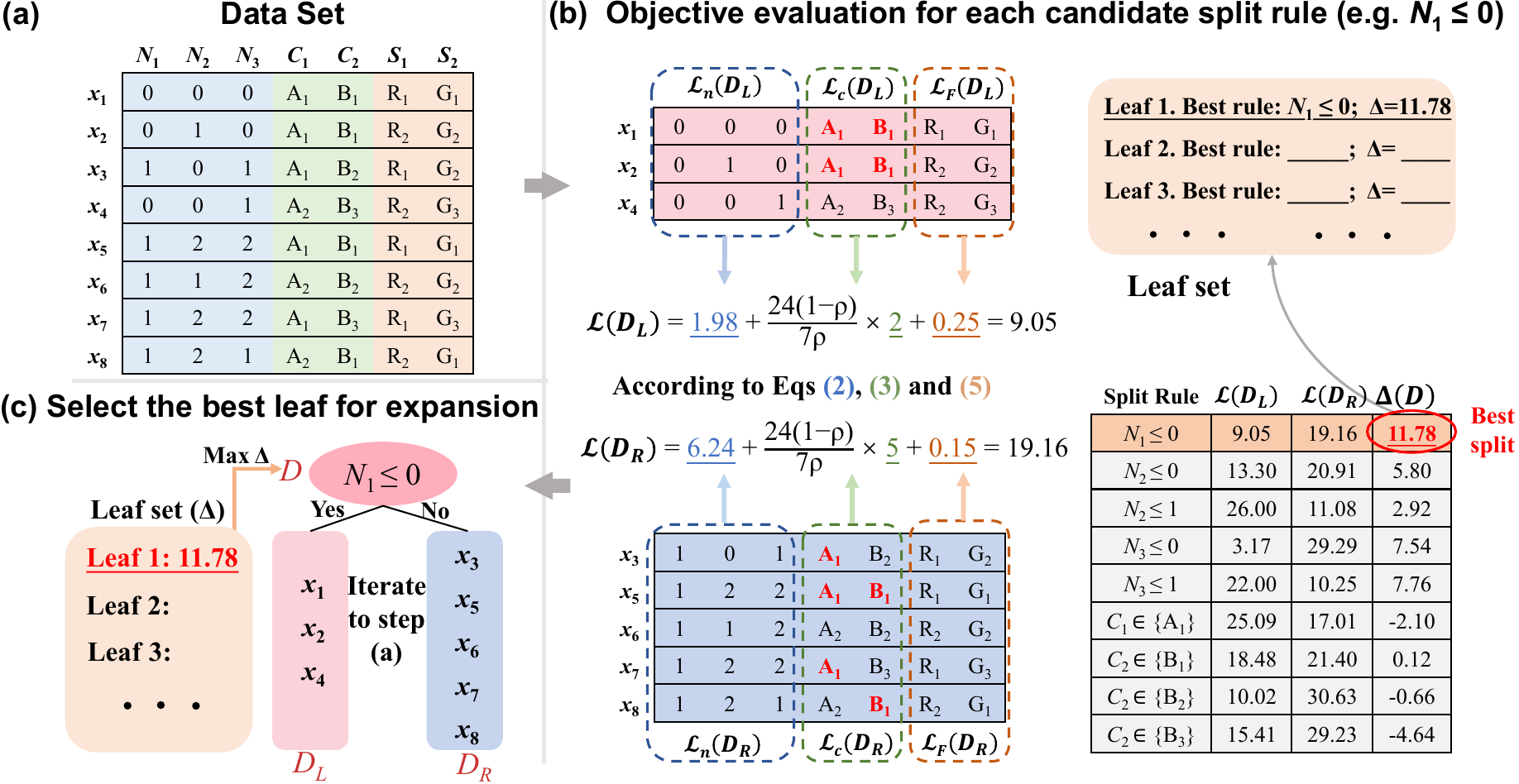}  
	\centering
	\caption{Illustration of the IFCT growth process: (a) A toy dataset comprising numerical attributes ($N_1,N_2,N_3$),  categorical attributes ($C_1,C_2$), and sensitive attributes ($S_1,S_2$). (b) Each candidate split rule is evaluated based on the objective function, with $N_1 \leq 0$ shown as an example. The dataset is split into $D_L$ and $D_R$, and the total loss  $\mathcal{L} (D_L)$ includes the compactness loss of numerical features $\mathcal{L}_{n} (D_L)$, categorical features $\mathcal{L}_{c} (D_L)$, and  fairness regularization $\mathcal{L}_{F} (D_L)$, computed according to Eqs. (2), (3), and (5), respectively. The loss of $D_R$ is computed similarly. Given $\rho=\frac{3}{2+3}=0.6$, the rule with the highest gain $\Delta (D)$ is selected and added to the leaf set. (c)  The leaf node with the maximum $\Delta$ is then selected for expansion using a best-first strategy, resulting in an updated tree.}  
	\label{fig1}   
\end{figure}

The algorithm begins by computing the global distribution of sensitive groups, which serves as the reference for fairness evaluation (line~1). It then initializes the tree $\mathcal{T}$ with a single root node $r$ containing all samples and defines the initial leaf set as $\mathbb{L} = \{r\}$ (lines~2–3).
For each leaf node $D$, a set of candidate split rules is evaluated to identify the one that maximizes the gain $\Delta(D)$ (line~4), defined as the reduction in the objective loss after the split:
$\Delta(D) = \mathcal{L}(D) - \big[\mathcal{L}(D_L) + \mathcal{L}(D_R)\big].$
For numerical features, candidate thresholds are sampled between adjacent distinct values, while for categorical features, the algorithm enumerates subsets of categories to construct binary partition rules.
During tree growth, IFCT repeatedly selects the leaf node $D$ with the largest gain $\Delta(D)$ (line~6) and splits it into two child nodes $(D_L, D_R)$ according to its best rule (lines~7–9). Each new node then updates its local statistics and recomputes the gain and associated split rule (lines~10–11). This best-first expansion proceeds iteratively until the number of leaves reaches the desired cluster count $k$. 
Finally, the algorithm outputs the constructed tree $\mathcal{T}$, where each leaf corresponds to an interpretable cluster represented by a conjunction of feature-based decision rules. A toy example of the IFCT growth process is shown in Fig. \ref{fig1}.

\textbf{Time complexity analysis.} IFCT grows to $k$ leaves by evaluating candidate splits at each internal node over all numerical and categorical features. For numerical features, sorting $n$ samples and scanning adjacent thresholds incurs a per-node cost of $\mathcal{O} (d_{n}n\log{n})$. Aggregated over the entire tree, and noting that the average depth is $\mathcal{O} (\log{k})$, the total cost becomes $\mathcal{O} (d_{n}n\log{n}\log{k})$. For categorical features, the number of binary partitions grows exponentially with the feature cardinality $r_j$, resulting in a per-node cost of $\mathcal{O} (\sum_{j=1}^{d_\text{cat}}2^{r_j})$. Since the number of internal nodes is proportional to $k$, the total cost for categorical features is $\mathcal{O} (k\sum_{j=1}^{d_\text{cat}}2^{r_j})$. Therefore, the overall computational complexity of IFCT is $\mathcal{O} (d_{n}n\log{n}\log{k}+k\sum_{j=1}^{d_\text{cat}}2^{r_j})$.

\subsection{Interpretable Fair Clustering Tree with Post-Pruning}
\begin{algorithm}[htbp]
\caption{IFCT-P}
\label{alg:iftree-p}
\begin{algorithmic}[1]
\Require Dataset $\mathcal{X}$, number of clusters $k$.
\Ensure Interpretable decision tree $\mathcal{T}$ with $k$ leaves.
\State Compute global sensitive group distribution $G$ 
\State Initialize tree $\mathcal{T}$ with root node $r \gets node(\mathcal{X})$
\State Initialize leaf set $\mathbb{L} \gets \{r\}$
\State Initialize candidate pruning set $\mathbb{C} \gets \emptyset$
\State $\Delta_{C}(r)$, $SR_r\gets\textsc{BestRule}(r)$
\While{$\exists\, v \in \mathbb{L} \text{ such that } SR_v \neq \varnothing$}
    \State $D \gets \arg\max_{v \in \mathbb{L}} \Delta_{C}(v)$
    \State $(D_L, D_R) \gets SR_{D}(D)$
    \If{$|D_L| < n_{min}$ \textbf{ or } $|D_R| < n_{min}$}
        \State $SR_D \gets \varnothing$
        \State $\Delta_{C}(D) \gets -\infty$
        \State \textbf{continue}
    \EndIf
    \State Attach $(D_L, D_R)$ as children of $D$ in $\mathcal{T}$
    \State $\mathbb{L} \gets (\mathbb{L} \setminus \{D\}) \cup \{D_L, D_R\}$
    \State $\mathbb{C} \gets \mathbb{C} \cup \{D\}$
    \State $\Delta_{C}(D_L), SR_{D_L} \gets \textsc{BestRule}(D_L)$
    \State $\Delta_{C}(D_R), SR_{D_R} \gets \textsc{BestRule}(D_R)$
\EndWhile
\While{$|\mathbb{L}| > k$}
  \State $D \gets \arg\max_{v \in \mathbb{C}} \Delta_F(v)$
  \While{$|\mathbb{L}| - |\mathbb{L}_D| + 1 < k$}
    \State $\mathbb{C} \gets \mathbb{C} \setminus \{D\}$
    \State $D \gets \arg\max_{v \in \mathbb{C}} \Delta_F(v)$
  \EndWhile
  \State $\mathbb{L} \gets \big(\mathbb{L} \setminus \mathbb{L}_D\big) \cup \{D\}$
  \State $\mathbb{C} \gets \mathbb{C} \setminus \mathbb{C}_D $
\EndWhile
\State \Return $\mathcal{T}$
\end{algorithmic}
\end{algorithm}

Although the proposed IFCT method takes both clustering compactness and fairness into account during tree construction, its hierarchical structure fixes early splits, limiting flexibility for global adjustments. Moreover, the trade-off coefficient $\lambda$ typically requires manual tuning to achieve a satisfactory balance. To address these limitations, we develop an enhanced variant, IFCT-P (Interpretable Fair Clustering Tree with Post-pruning), which first grows the tree with a focus on clustering compactness and then applies a fairness-guided post-pruning strategy to refine the structure.

IFCT-P follows a two-stage procedure, as outlined in Algorithm~\ref{alg:iftree-p}. In the first stage, the algorithm constructs an over-expanded tree to ensure sufficient structural flexibility for later pruning. This process mirrors that of IFCT, except that fairness constraints are omitted (lines~6-19). Specifically, the splitting gain $\Delta_{C}(D)$ is computed solely based on compactness loss $\Delta_{C}(D) = \mathcal{L}_{C}(D) - \big[\mathcal{L}_{C}(D_L) + \mathcal{L}_{C}(D_R)\big]$, allowing the tree to grow purely according to structural compactness. The expansion proceeds until all leaf nodes contain fewer than $n_{min}$ instances (which is set to 1 in practice to avoid manual tuning) (lines~9-13). 

In the second stage, the algorithm iteratively prunes subtrees that contribute least to the fairness-regularized objective. Let $\mathbb{L}_{D}$ denote the set of leaf nodes in the subtree rooted at node $D$, and $\mathbb{C}_{D}$ the set of its internal nodes. For each internal node $D$, the pruning gain $\Delta_F(D)$ is defined as the reduction in fairness-adjusted loss obtained by replacing the entire subtree with a single merged leaf:
\begin{equation}
    \Delta_F(D)=\frac{\sum_{v \in \mathbb{L}_{D}} \mathcal{L}_{F}(v)}{|\mathbb{L}_{D}|}-\mathcal{L}_{F}(D).
\end{equation}
The node with the highest $\Delta_F(D)$ is selected for pruning. To prevent over-pruning, any candidate node whose removal would reduce the total number of leaves below $k$ is excluded from the candidate set $\mathbb{C}$ and the next best candidate is considered instead (lines~22-25). This constraint ensures that the pruning process terminates with exactly $k$ interpretable clusters, yielding the final clustering structure.

\textbf{Time complexity analysis.} Compared to IFCT, IFCT-P exhibits similar time complexity, with the main distinction arising from its over-expansion and post-pruning strategy. When the tree is fully expanded such that each leaf contains a single sample, the overall complexity becomes $\mathcal{O} (d_{n}n\log^{2}{n}+k\sum_{j=1}^{d_\text{cat}}2^{r_j})$. The post-pruning phase incurs negligible overhead, as fairness-regularized gains are precomputed during  tree growth, and each pruning step requires only a heap update without recomputing statistics.

\section{Experiments}
\label{4}
This section presents a series of experiments designed to evaluate the performance of the proposed method. All experiments were conducted on a workstation configured with an Intel(R) Core(TM) i7-10700F processor running at 2.90 GHz, 16 GB of RAM, and an NVIDIA GeForce RTX 1660 GPU with 6 GB of dedicated memory.

\subsection{Datasets}
The proposed method was evaluated on five benchmark fairness datasets, including three real-world datasets: Bank Marketing \cite{moro2014data}, Credit Card \cite{zheng2023fairness}, and HCV \footnote{https://archive.ics.uci.edu/dataset/571/hcv+data}, which are widely used in fair clustering studies, and two synthetic datasets: 2d-4c-no1 and 2d-10c-no1 \footnote{https://personalpages.manchester.ac.uk/staff/Julia.Handl/data.tar.gz}, which are commonly adopted in clustering research. Since the synthetic datasets lack predefined sensitive attributes, protected group labels are randomly generated using a Bernoulli distribution with $p=0.5$. The detailed characteristics of each dataset, including the number of samples, features, clusters, and sensitive attributes, are summarized in Table \ref{dataset}.

\begin{table}[htbp]
	\caption{The summary statistics on datasets used in the performance evaluation.}
	\label{dataset}
	\centering
        \setlength{\tabcolsep}{0.62mm}{
	\begin{tabular}{ccccccc}
		\toprule
		Dataset &  \#Samples &\#Numerical Features&\#Categorical Features&  \#Clusters &Target attribute & Sensitive Feature \\
		\midrule
		Bank Marketing &  4108  &10 &9 &  2   &  Subscription Status  &   Marital      \\
            HCV &  2907 &11 &0  &  5   &  Diagnostic Category   &   Gender      \\
            Credit Card &  30000 &14 &8  &  2       &  Default Payment   &  Gender   \\
		2d-4c-no1 &  1623  &2 & 0&  4   &  Synthetic Label   &   Synthetic Binary    \\
		2d-10c-no1 &  2525  &2 &0 &  10      &  Synthetic Label  &  Synthetic Binary  \\
		\bottomrule
	\end{tabular}}
\end{table}

\subsection{Experimental Setup}

\subsubsection{Comparision Methods}
The following clustering algorithms are used for comparison with our method and can be categorized into two groups.

\textit{Fair clustering methods}:
\begin{itemize}
    \item FFC \cite{pan2023fairness}: A multi-stage approach that enforces fairness during initialization, relaxes fairness constraints to enhance clustering quality, and refines the results through a fairness-preserving local search procedure.
    \item BFKM \cite{pan2023balanced}: This method integrates fairness and balance constraints into the $k$-means objective by penalizing deviations in group representation and cluster sizes, and solves the optimization problem using coordinate descent.
    \item VFC \cite{ziko2021variational}: This method adopts a variational formulation that incorporates a KL-divergence-based fairness penalty into various clustering objectives, enabling a flexible trade-off between clustering quality and demographic balance through a unified bound optimization framework.
    \item FairSC \cite{pmlr-v97-kleindessner19b}: A spectral clustering method that guarantees individual and group fairness by constructing a fair similarity graph and solving a constrained eigenvalue problem to obtain balanced cluster assignments.
    \item FairDen \cite{krieger2025fairden}: A fair density-based clustering method that captures density-connectivity in a continuous form, supports categorical features and multiple sensitive attributes, and enforces fairness through spectral optimization.
\end{itemize}

\textit{Interpretable clustering methods}:
\begin{itemize}
    \item IMM \cite{dasgupta2020explainable}: This method builds a threshold decision tree with 
$k$ leaves, where $k$ corresponds to the number of ground-truth clusters, by minimizing the number of classification errors at each node. It achieves an approximation ratio comparable to that of the  $k$-means or $k$-medians objectives.
    \item ExKMC \cite{frost2020exkmc}: This approach constructs an initial threshold tree with $k$ leaves and then expands it to up to $k'$ leaves, where $k'$ is a user-defined parameter satisfying $k' > k$. In our implementation, we set $k'=2k$, with $k$ denoting the number of ground-truth clusters.
    \item Shallow \cite{laber2023shallow}: This method optimizes the $k$-means objective while introducing a regularization term that penalizes tree depth, promoting the construction of shallow and interpretable decision trees.
\end{itemize}

For datasets containing both numerical and categorical features (i.e., Bank Marketing and Credit Card), only the numerical features are retained as input for all baseline methods except FairDen, as these methods cannot handle mixed-type data. The parameters of all baseline methods are configured according to their original papers or default settings provided in their official implementations. For our methods, the balance parameter $\lambda$ in IFCT is set to $10^4$, while IFCT-P requires no additional parameter tuning.

\subsubsection{Evaluation Measures}
We evaluate the clustering performance of the proposed method using Clustering Accuracy (ACC) and Normalized Mutual Information (NMI), two widely adopted metrics for assessing the alignment between predicted cluster assignments and ground-truth labels. In both cases, higher values indicate better clustering quality.

To quantitatively evaluate group fairness in clustering results, we adopt two widely used metrics  Balance (BAL) \cite{zheng2023fairness} and Minimum Normalized Cluster Entropy (MNCE) \cite{zeng2023deep}, defined as follows:
\begin{equation}
\text{BAL} = \min_i \left( \frac{\min_j |\Omega_i \cap G_j|}{|\Omega_i|} \right),
\end{equation}
and
\begin{equation}
\text{MNCE} = 
\frac{
\displaystyle \min_i \left( -\sum_j \frac{|G_j \cap \Omega_i|}{|\Omega_i|} \log \frac{|G_j \cap \Omega_i|}{|\Omega_i|} \right)
}{
\displaystyle -\sum_j \frac{|G_j|}{n} \log \frac{|G_j|}{n}
},
\end{equation}
respectively. Here, $\Omega_i$ denotes the set of instances assigned to the $i$-th cluster, and 
$G_j$ represents the set of instances belonging to the $j$-th sensitive group. In both metrics, higher values indicate better group fairness across clusters.

\subsection{Experimental Results}
\begin{table}[htbp]
\setlength{\tabcolsep}{5pt}
	\caption{Comparison of various clustering methods in terms of two accuracy metrics and two fairness metrics. For each metric across the datasets, the highest score among all methods is shown in bold, while the best-performing fair clustering method is additionally underlined. A “–” in the table indicates that FairDen and FairSC failed to produce clustering results due to excessive memory consumption on large-scale datasets.}
	\label{compare}
	\centering
		\begin{tabular}{c|cccc|ccccccc}
			
			\toprule
			Dataset &Metrics&IMM&ExKMC&Shallow&FFC&BFKM&VFC&FairDen&FairSC&IFCT&IFCT-P\\
			\midrule
			&   ACC    &\textbf{0.900} &\textbf{0.900}   &0.859 &0.633 &0.636 &0.647  &0.682 &0.535 &0.718 &\underline{0.719} \\
			Bank Marketing &  NMI  &\textbf{0.126}&\textbf{0.126}&0.129&0.051&0.051&0.057&\underline{0.092}&0&0.073&0.076 \\
			&   BAL   &0.075&0.075&0.104&0.107&\textbf{0.108}&0.102&0.103&0.002&0.100&0.105  \\
            &   MNCE   &0.991&0.991&0.996&0.978&0.981&0.981&0.98&\textbf{0.998}&0.979&0.991 \\
			\midrule
                &   ACC    &0.829&\textbf{0.842}&0.831&0.281&0.262&0.293&0.368&0.350 &0.367&\underline{0.424} \\
			HCV &    NMI    &0.169&\textbf{0.288}&0.185&0.150&0.085&0.101&0.230&0.093&0.175&\underline{0.251}  \\
			&   BAL   &0&0&0&0.310&0.347&0.375&0.200&0.359&\textbf{0.386}&0.263  \\
            &   MNCE   &0&0&0&0.928&0.967&0.991&0.750&0.978&\textbf{0.998}&0.864 \\
                \midrule
                &   ACC    &\textbf{0.703}&0.697&0.701&0.594&0.594&0.541&-&-&0.698&\textbf{0.703} \\
			Credit Card &    NMI    &0&0.001&0&\textbf{0.002}&\textbf{0.002}&0&-&-&0.001&0.001 \\
			&   BAL   &0.392&0.391&0.392&0.393&0.393&0.386&-&-&\textbf{0.394}&0.393  \\
            &   MNCE   &0.997&0.997&0.997&0.998&0.998&0.993&-&-&\textbf{0.999}&0.998  \\
			\midrule
			&   ACC    &0.977&0.721&0.912&0.862&0.743&0.836&\textbf{0.988}&0.624&0.776&0.713 \\
			2d-4c-no1 &    NMI    &0.805&0.795&0.811&0.787&0.762&0.744&\textbf{0.949}&0.262&0.607&0.761  \\
			&   BAL   &0.439&0.438&0.438&0.442&0.443&0.463&0.444&0.460 &\textbf{0.485}&0.440  \\
            &   MNCE   &0.989&0.990&0.989&0.991&0.991&0.997&0.992&0.996&\textbf{1}&0.990  \\
			\midrule
			&   ACC    &0.783&0.780&0.777&0.728&0.571&0.704&\textbf{0.970}&0.561&0.545&0.531 \\
			2d-10c-no1 &    NMI    &0.824&0.826&0.824&0.738&0.684&0.715&\textbf{0.937}&0.318&0.539&0.621 \\
			&   BAL   &0.397&0.408&0.408&0.409&0.427&0.467&0.405&0.442&\textbf{0.480}&0.444  \\
            &   MNCE   &0.969&0.975&0.976&0.976&0.985&0.997&0.974&0.990&\textbf{0.999}&0.991  \\
            
			\bottomrule
	\end{tabular}
\end{table}

Table \ref{compare} presents the detailed comparison results, from which we derive the following observations and insights.

\underline{Compared with interpretable clustering methods}: In terms of the two accuracy metrics, ACC and NMI, our methods (IFCT and IFCT-P) generally perform worse than the interpretable clustering methods IMM, ExKMC, and Shallow on most datasets, except for the Credit Card dataset. This result highlights the inherent trade-off between clustering accuracy and fairness—improving fairness often comes at the cost of reduced clustering performance.
However, with respect to the two fairness metrics, BAL and MNCE, IFCT and IFCT-P consistently exhibit clear advantages across both real-world and synthetic datasets, demonstrating their effectiveness in promoting group fairness.

\underline{Compared with fair clustering methods}: In terms of fairness, IFCT outperforms all other fair clustering methods on all datasets except Bank Marketing, while IFCT-P also demonstrates strong competitiveness on the Bank Marketing, Credit Card, and 2d-4c-no1 datasets.
Regarding accuracy, although both IFCT and IFCT-P underperform on the two synthetic datasets with purely numerical features, IFCT-P achieves the highest ACC scores on all three real-world datasets. This result suggests that our method is particularly effective in handling mixed-type data.

\underline{Overall performance}: As previously discussed, our proposed methods demonstrate strong performance in both fairness and accuracy while maintaining interpretability. These results confirm the effectiveness of our approach in addressing fair clustering tasks.
Moreover, when comparing IFCT and IFCT-P, we observe that IFCT consistently achieves better fairness, whereas IFCT-P tends to outperform IFCT in terms of NMI. This contrast can be attributed to two main factors. First, IFCT adopts a relatively large balance parameter $\lambda$, which encourages more globally balanced cluster assignments, potentially at the cost of some clustering accuracy. Second, IFCT-P does not incorporate fairness considerations during the initial decision tree construction phase, which may result in tree structures more aligned with accuracy rather than fairness.

\subsection{Parameter Sensitivity}
\begin{figure}[pos=H] 
	\centering  
	\vspace{-0.05cm} 
	\subfigtopskip=2pt 
    \subfigbottomskip=2pt 
	\subfigcapskip=-5pt 
	\subfigure[HCV]{
		\label{hcv}
		\includegraphics[width=0.45\linewidth]{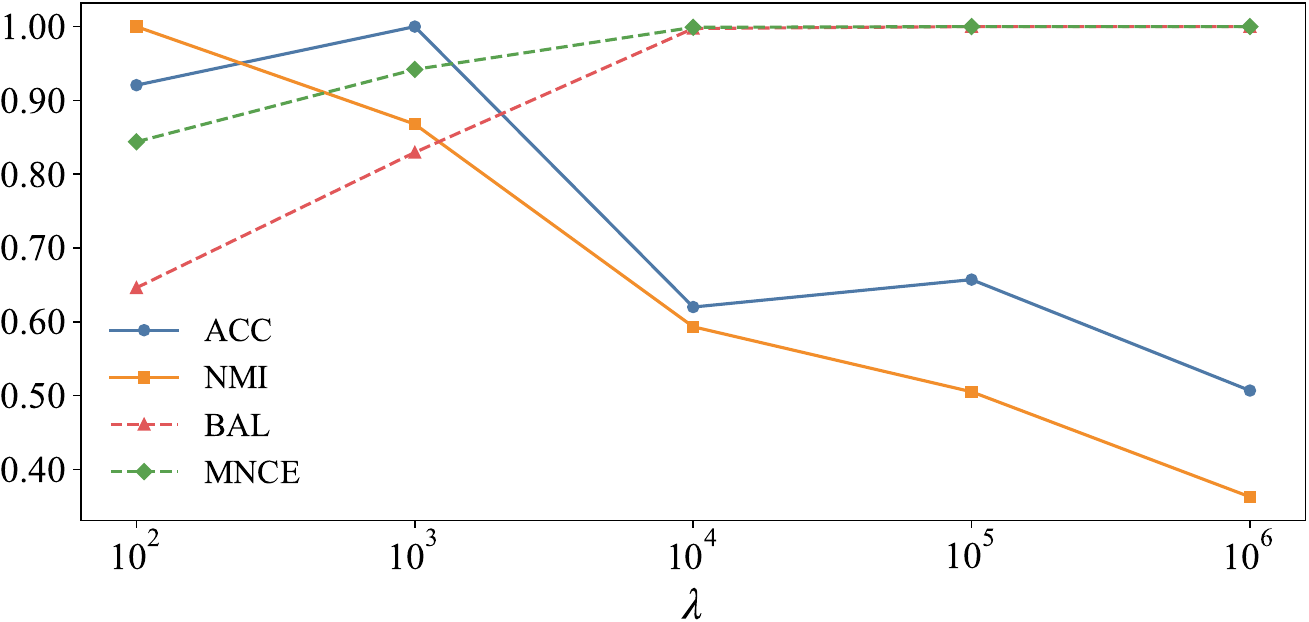}}
	\quad 
	\subfigure[2d-10c-no1]{
		\label{2d}
		\includegraphics[width=0.43\linewidth]{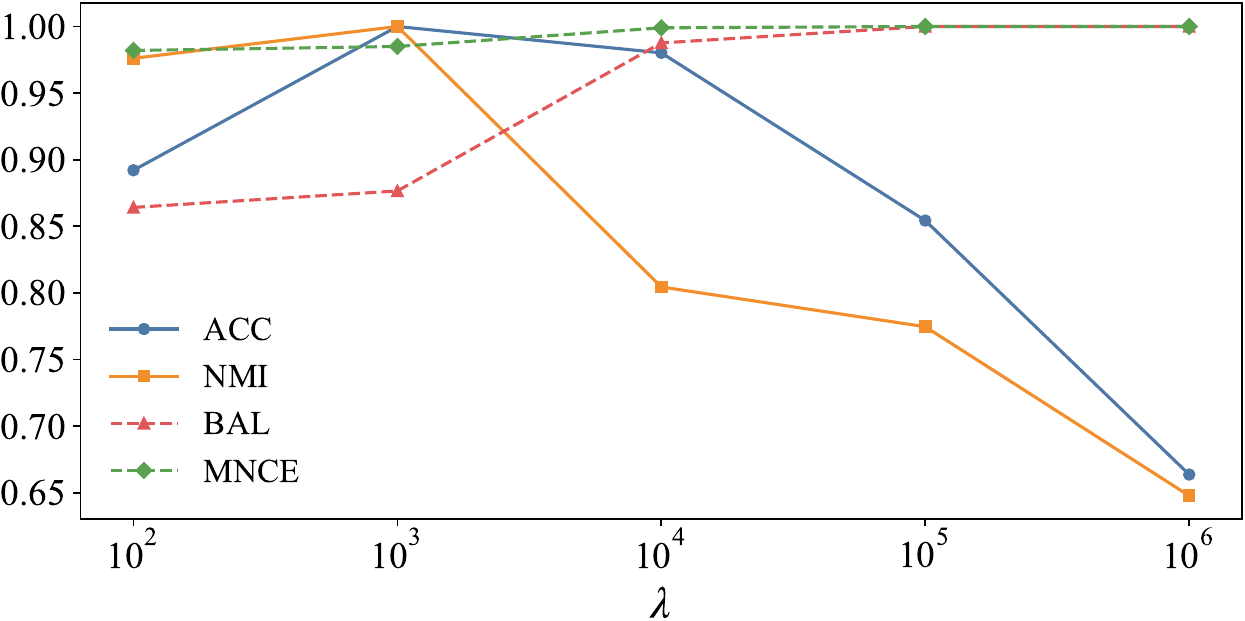}}
	\caption{Effect of the parameter $\lambda$ on all evaluation metrics for IFCT. Each metric is normalized to the range $[0,1]$ by dividing it by its maximum value across all $\lambda$ settings.}
	\label{sen}
\end{figure}

In IFCT, the trade-off parameter $\lambda$ plays a critical role in balancing clustering accuracy and fairness. To examine its influence, we vary $\lambda$ from $10^2$ to $10^6$ and observe the resulting variations across the four evaluation metrics.

As shown in Fig. \ref{sen}, all metrics are notably sensitive to the choice of $\lambda$. When 
$\lambda$ increases, the decision tree construction is increasingly guided by fairness considerations during splits. As a result, both fairness metrics (BAL and MNCE) show clear improvements, while the accuracy metrics (ACC and NMI) tend to decline significantly. This illustrates the inherent trade-off between accuracy and fairness during the growth of the IFCT.

In contrast, IFCT-P does not involve any tunable hyper-parameters. This not only reduces the need for manual tuning but also ensures stable performance without the need for parameter tuning, making it more robust across different datasets and settings.
\subsection{Decision Tree Visualization}
\begin{figure}[pos=H] 
	\centering  
	\vspace{-0.05cm} 
	\subfigtopskip=2pt 
    \subfigbottomskip=2pt 
	\subfigcapskip=-5pt 
	\subfigure[IFCT]{
		\label{IFCT}
		\includegraphics[width=0.3\linewidth]{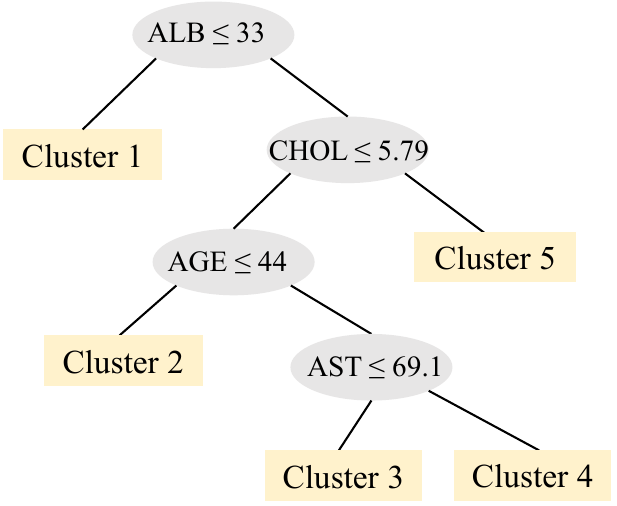}}
	\quad 
	\subfigure[IFCT-P]{
		\label{IFCT-P}
		\includegraphics[width=0.3\linewidth]{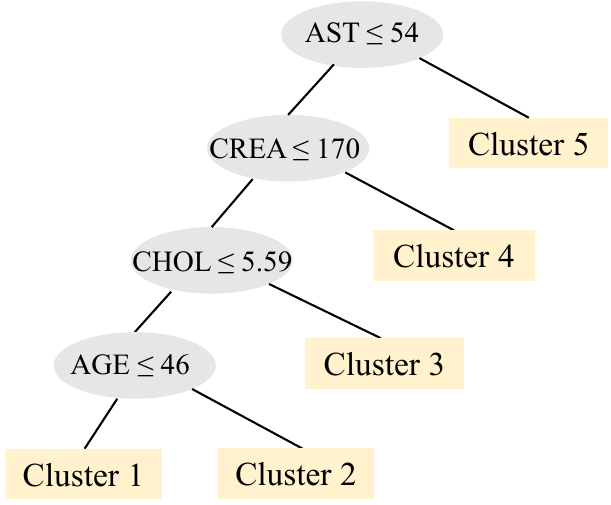}}
	\caption{Visualization of decision trees constructed by IFCT and IFCT-P on the HCV dataset.}
	\label{tree}
\end{figure}

To illustrate the interpretability of our proposed methods, Fig.~\ref{tree} presents the decision trees generated by IFCT and IFCT-P on the HCV dataset. Both methods yield compact trees with a limited number of splits, enabling intuitive understanding of the clustering process. These visualizations demonstrate that the cluster assignments produced by our models can be transparently traced through feature-based decisions—an essential advantage in applications where model behavior must be explainable to human users.
\subsection{Multiple Sensitive Attributes}

To evaluate the ability of our method to effectively handle multiple sensitive attributes, we conduct experiments on the Credit Card dataset. As FairDen is the only baseline capable of processing multiple sensitive attributes, we include it for comparison with our IFCT method. However, due to FairDen’s excessive memory consumption on large-scale data, we randomly sample 5000 instances from the dataset to ensure that it can run successfully. In this experiment, we jointly consider three sensitive attributes commonly used in fairness studies: Gender (G) (male/female), Education (E) (graduate school/university/high school), and Marital Status (M) (married/single).
\begin{table}[pos=H]
\centering
\caption{BAL scores of FairDen and IFCT on the Credit Card dataset under various sensitive attribute configurations. For each setting, fairness is evaluated with respect to each target attribute, and the higher score  is shown in bold.}
\begin{tabular}{ccc c}
\toprule
Sensitive Attributes & Target & FairDen & IFCT \\
\midrule
G            & G       & 0.395  & \textbf{0.396} \\
E            & E     & \textbf{0.139}  & 0.116 \\
M            & M         & 0.452  & \textbf{0.457} \\
\midrule
             & G         & 0.392  & 0.394 \\
G \& E        & E      & 0.098  & 0.115 \\
             & Average            & 0.245  & \textbf{0.255} \\
\midrule
             & G        & 0.392  & 0.389 \\
 G \& M       & M         & 0.453  & 0.456 \\
             & Average            & \textbf{0.423}  & \textbf{0.423} \\
\midrule
             & E      & 0.090  & 0.115 \\
   E \& M     & M         & 0.449  & 0.456 \\
             & Average            & 0.270  & \textbf{0.286} \\
\midrule
             & G        & 0.388  & 0.393 \\
G \& E \& M    & E      & 0.005  & 0.115 \\
             & M         & 0.428  & 0.456 \\
             & Average            & 0.274  & \textbf{0.315} \\
\bottomrule
\end{tabular}
\label{multi}
\end{table}

From Table~\ref{multi}, we observe that IFCT consistently achieves comparable or superior fairness performance across most sensitive attribute configurations. Specifically, when considering individual attributes such as Gender (G), Education (E), and Marital Status (M), IFCT yields the highest BAL scores in two out of three cases.
For combinations of multiple sensitive attributes, IFCT demonstrates clear advantages. In the settings of {G \& E}, {E \& M}, and {G \& E \& M}, the average BAL achieved by IFCT surpasses that of FairDen by a notable margin. These results indicate that IFCT is capable of effectively balancing fairness across multiple sensitive dimensions, confirming its robustness in handling multi-attribute fairness and promoting equitable clustering outcomes.

\subsection{Running Time Comparison}
\begin{figure}[pos=H] 

	\centering  
	\vspace{-0.05cm} 
	\subfigtopskip=2pt 
	\subfigbottomskip=2pt 
	\subfigcapskip=-5pt 
	\subfigure[Bank Marketing]{
		\includegraphics[width=0.17\linewidth]{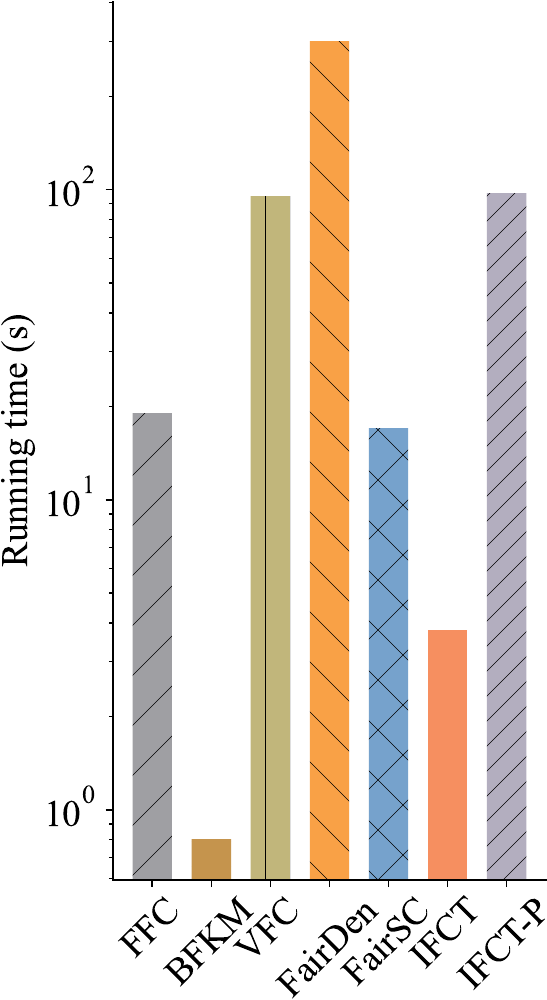}}
	\quad 
	\subfigure[HCV]{
		\includegraphics[width=0.17\linewidth]{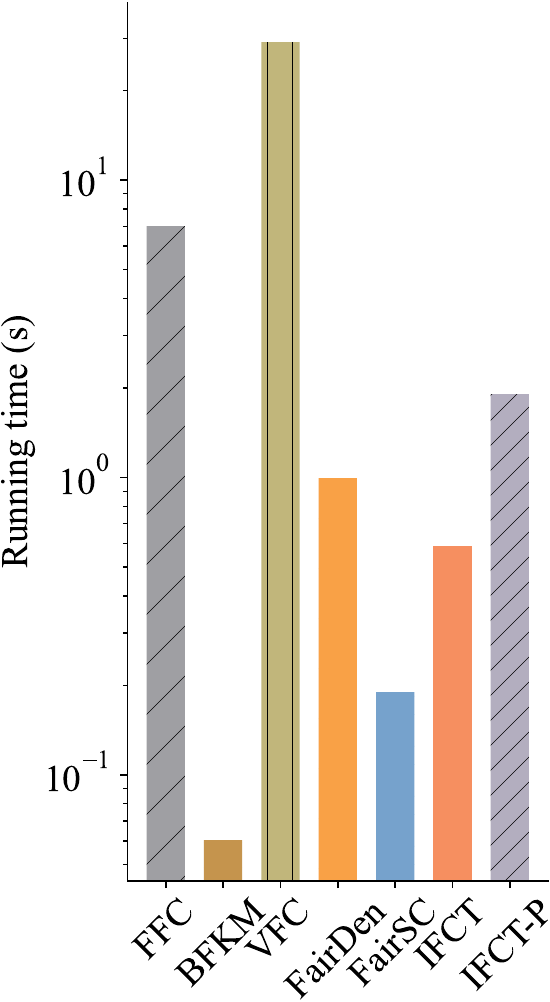}}
        \quad 
	\subfigure[Credit Card]{
		\includegraphics[width=0.17\linewidth]{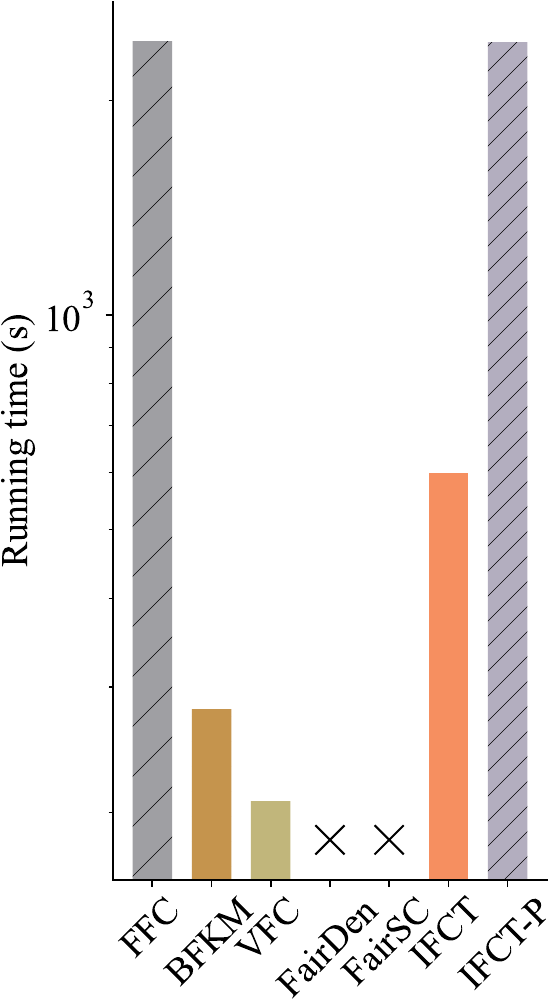}}
        \quad 
	\subfigure[2d-4c-no1]{
		\includegraphics[width=0.17\linewidth]{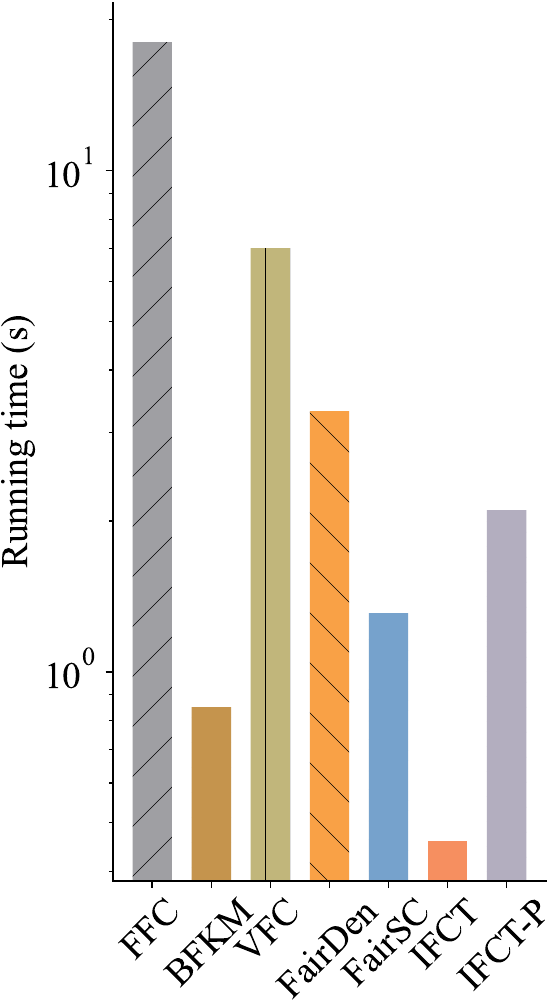}}
	   \quad 
	\subfigure[2d-10c-no1]{
		\includegraphics[width=0.17\linewidth]{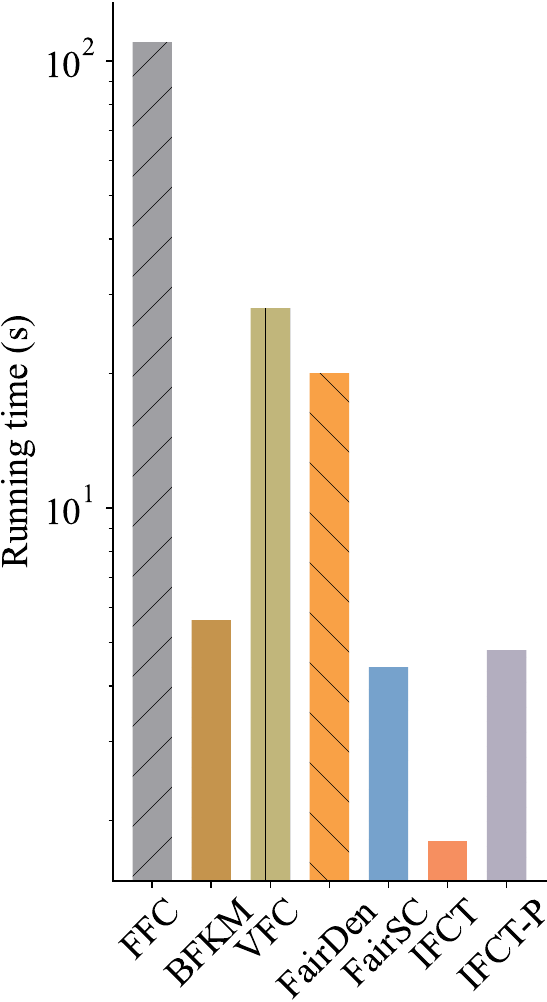}}

	\caption{Running time comparison  across different datasets.}
 \label{time}
\end{figure}

Fig. \ref{time} presents the runtime comparison between our proposed methods and other fair clustering baselines. As shown in Fig. \ref{time}, IFCT consistently ranks among the top three in terms of running time across all datasets, and achieves the best performance on the two low-dimensional synthetic datasets. In contrast, IFCT-P exhibits higher sensitivity to dataset size due to the construction of oversized trees in the initial phase, which are later pruned to satisfy fairness constraints. Despite this, its overall runtime remains within a moderate range compared to other baselines.
\section{Conclusion}
\label{5}
In this paper, we present IFCT, an interpretable and fair clustering framework based on decision trees. By explicitly incorporating fairness constraints into the tree construction process, our method provides transparent decision boundaries while ensuring balanced treatment across protected groups. We also introduce a post-pruning variant that eliminates the need for fairness-related hyperparameters. Experimental results on both real-world and synthetic datasets demonstrate that our method achieves competitive clustering performance, superior group fairness, and strong interpretability, even under complex fairness settings involving multiple sensitive attributes.

For future work, we plan to explore alternative interpretable clustering models, such as \textit{if-then} rules or prototype-based approaches, to better accommodate complex data structures while preserving model transparency and fairness. In addition, more flexible fairness formulations may be explored to effectively handle overlapping or hierarchical sensitive attributes.

\section*{Acknowledgments}
This work has been supported by the Science and Technology Planning Project of Liaoning Province under Grant No. 2023JH26/10100008, and the  National Natural Science Foundation of China under Grant Nos. 62476038, and 62472064. 
\bibliographystyle{cas-model2-names}

\bibliography{cas-refs}



\end{document}